%% file: main.tex
\documentclass[letterpaper]{article} 
\usepackage{aaai25}  
\usepackage{times}  
\usepackage{helvet}  
\usepackage{courier}  
\usepackage[hyphens]{url}  
\usepackage{graphicx} 
\usepackage{natbib}  
\usepackage{caption} 
\usepackage{algorithm}
\usepackage{algorithmic}
\usepackage{tabularray}
\UseTblrLibrary{siunitx}
\usepackage{tikz}
\usepackage{rotating}
\usepackage{fontawesome5}
\usepackage{newfloat}
\usepackage{listings}
\DeclareCaptionStyle{ruled}{labelfont=normalfont,labelsep=colon,strut=off} 
\floatstyle{ruled}
\newfloat{listing}{tb}{lst}{}
\floatname{listing}{Listing}
\title{Enhancing Portuguese Variety Identification with Cross-Domain Approaches}
\author {
    Hugo Sousa\textsuperscript{\rm 1,\rm 3}\equalcontrib,
    Rúben Almeida\textsuperscript{\rm 2,\rm 3,\rm 4}\equalcontrib,
    Purificação Silvano\textsuperscript{\rm 5,\rm 6}, \\
    Inês Cantante\textsuperscript{\rm 5,\rm 6},
    Ricardo Campos\textsuperscript{\rm 3, \rm 7,\rm 8},
    Alípio Jorge\textsuperscript{\rm 1,\rm 3}
}
\affiliations {
    \textsuperscript{\rm 1}Faculty of Sciences, University of Porto, Porto, Portugal\\
    \textsuperscript{\rm 2}Faculty of Engineering, University of Porto, Porto, Portugal\\
    \textsuperscript{\rm 3}INESC TEC, Porto, Portugal\\
    \textsuperscript{\rm 4}Innovation Point - dst group, Braga, Portugal\\
    \textsuperscript{\rm 5}Faculty of Arts and Humanities, University of Porto, Porto, Portugal\\
    \textsuperscript{\rm 6}Centre of Linguistics, University of Porto, Porto, Portugal \\
    \textsuperscript{\rm 7}Department of Informatics, University of Beira Interior, Covilhã, Portugal \\
    \textsuperscript{\rm 8}Ci2 - Smart Cities Research Center, Tomar, Portugal \\
    hugo.o.sousa@inesctec.pt, ruben.f.almeida@inesctec.pt
}

\begin{document}

\maketitle

\input{src/Abstract}
\input{src/Introduction}

\input{src/Related_Work}
\input{src/Dataset}

\input{src/Experimental_Setup}
\input{src/Results}

\input{src/Conclusion}

\input{src/Limitations_and_Ethical}
\input{src/aknowledments}

\bibliography{references}

\end{document}

%% file: src/Abstract.tex
\begin{abstract}

Recent advances in natural language processing have raised expectations for generative models to produce coherent text across diverse language varieties. In the particular case of the Portuguese language, the predominance of Brazilian Portuguese corpora online introduces linguistic biases in these models, limiting their applicability outside of Brazil. To address this gap and promote the creation of European Portuguese resources, we developed a cross-domain language variety identifier (LVI) to discriminate between European and Brazilian Portuguese. Motivated by the findings of our literature review, we compiled the PtBrVarId corpus, a cross-domain LVI dataset, and study the effectiveness of transformer-based LVI classifiers for cross-domain scenarios. Although this research focuses on two Portuguese varieties, our contribution can be extended to other varieties and languages. We open source the code, corpus, and models to foster further research in this task.

\end{abstract}

%% file: src/Introduction.tex
\section{Introduction}
\label{sec:introduction}

Discriminating between varieties of a given language is an important natural language processing (NLP) task~\cite{joshi2024natural}. Over time, populations that share a common language can evolve distinctive speech traits due to geographical and cultural factors, including migration and the influence of other languages~\cite{Raposo_Vicente_Veloso_2021}. Recently, this importance became even more pronounced with the advent of variety-specific large language models, where variety discrimination plays a pivotal role~\cite{Rodrigues_2023}. Whether in the pre-training, fine-tuning, or evaluation phase, having a highly effective system to discriminate between varieties reduces the amount of human supervision required, accelerating the production of curated mono-variety datasets~\cite{ohman2023nordic}. However, developing such a system presents considerable challenges. Classifiers often struggle to identify linguistically relevant features, showing a tendency to be biased towards non-linguistic factors, such as named entities and thematic content~\cite{diwersy2014weakly}. Consequently, these classifiers exhibit limited transfer capabilities to domains not represented in the training set, significantly restricting their utility in multi-domain applications~\cite{lui2011cross, nguyen-etal-2021-cross-domain}.

A language in which variety identification is particularly challenging is Portuguese. It is spoken by more than $200$ million people worldwide and serves as the official language of eight nations on five continents, each with its own variety. However, $70\%$ of Portuguese speakers are Brazilian citizens\footnote{Statistic inferred form Wikipedia \url{https://en.wikipedia.org/wiki/Portuguese_language#Lusophone_countries}}, which implies that resources labeled as Portuguese are dominated by this language variety. Another important characteristic of Portuguese is that, unlike languages where differences are predominantly phonological, such as those of the North Germanic family~\cite{Holmberg2008}, the widespread of Portuguese has fostered considerable phonological, morphological, lexical, syntactic and semantic variations between Portuguese varieties~\cite{Brito2016}. In the development of language models, for example, this variety divergence has practical implications; models trained in Brazilian Portuguese generate texts that are markedly distinct from those trained in other Portuguese varieties~\cite{Rodrigues_2023}. This fact restrains the adoption of these models outside of Brazil in domains where formal non-Brazilian text is required, as is the case of legal and medical applications. This underscores the practical importance of developing effective LVI systems that can be deployed in production.

In this study, we describe the development of a cross-domain LVI classifier that discriminates between Brazilian and European Portuguese. To accomplish that, we start with a comprehensive listing of Portuguese LVI resources. The lack of multi-domain corpora motivated us to compile our own dataset. This corpus was then used in the development of our LVI classifier. For the training procedure we devised a training protocol that takes into account the cross-domain capabilities of models during evaluation. Furthermore, we also study the impact of masking named entities and thematic content embedded in the training corpus, a process named delexicalization~\cite{lui2014exploring}. To summarize, the contributions of this work are the following:

\begin{enumerate}

\item We introduce a novel cross-domain, silver-labeled LVI corpus for Brazilian and European Portuguese, compiled from open-license datasets;

\item We examine the impact of different levels of delexicalization on the overall effectiveness of LVI models;

\item We propose a training protocol for LVI models that yields better generalization performance;

\item We release the first open-source Portuguese LVI model, providing a valuable resource for future research and practical applications.
\end{enumerate}

The remainder of this paper is organized as follows: Section~\ref{sec:related_work} offers a comprehensive literature review on the state-of-the-art in Portuguese LVI. In Section~\ref{sec:create_corpus}, we introduce our compiled dataset, PtBrVarId, and present relevant statistics along with the results of a manual evaluation of the quality of the dataset. Section~\ref{sec:setup} describes the training protocol and the models developed, including the baselines and benchmarks used for comparison. The results are presented in Section~\ref{sec:results}, followed by a discussion of future research directions in Section~\ref{sec:conclusions}. 


%% file: src/Related_Work.tex
\section{Related Work}
\label{sec:related_work}

\subsection{Corpora}

Despite the numerous works developed in the LVI task, the first gold-labeled dataset that includes Portuguese corpora, the DSL-TL corpus~\cite{zampieri2023language}, was only introduced in 2023. This dataset used crowdsourcing to annotate approximately 5k Portuguese documents. The corpus are not only labeled as ``European'' and ``Brazilian'' Portuguese, but also a special ``Both or Neither'' label to signal those documents with insufficient linguistic marks to be considered part of one of these varieties. 

Prior to the release of this dataset, the evaluation process was often performed in silver-labeled data, collected using domain-specific heuristics. For instance, in the journalistic domain, it is common to assume the language variety of a document based on the newspaper origin; Brazilian newspapers' articles are assigned a Brazilian Portuguese label, while Portuguese ones are assigned a European Portuguese label~\cite{da2006identification, zampieri2012automatic, tan2014merging}. In the social media domain, a similar approach is frequently used.~\citet{castro2016discriminating} used geographic metadata collected on Twitter to assign a language variety to each document based on the authors location. Unfortunately, many of these Portuguese LVI resources are no longer available online. This limitation motivated us to collect and open-source our training data.

\subsection{Modeling Approaches}
\label{subsec:techniques_used}

The high effectiveness of N-gram-based systems observed in language identification studies~\cite{mcnamee2005language, martins2005language, chew2009optimizing}, a task closely related to LVI, motivated the application of these methods in the context of LVI. To this day, this approaches are still employed, with several submissions to the VarDial workshop\footnote{\url{https://aclanthology.org/venues/vardial/}} -- which compiles most of the recent studies in the LVI task -- achieving high effectiveness. Notable examples include Italian with an $F_1$ score of 0.90~\cite{jauhiainen-etal-2022-italian}, Uralic with an $F_1$ score of 0.94~\cite{bernier-colborne-etal-2021-n}, and Mandarin with an $F_1$ score of 0.91~\cite{yang-xiang-2019-naive}.

The adoption of transformer-based techniques~\cite{vaswani2017attention} in LVI has not been as rapid as in other NLP tasks. Recently, some studies have leveraged monolingual BERT-based models to fine-tune LVI classifiers for Romanian~\cite{zaharia-etal-2020-exploring} and French~\cite{bernier2022transfer}. However, in none of these cases were transformers capable of outperforming N-gram-based techniques, only achieving a $F_1$ score of $0.65$ in Romanian and $0.43$ in French. Similar results have been reported for different languages using other deep learning techniques, such as multilingual transformers~\cite{popa-stefanescu-2020-applying}, feedforward neural networks~\cite{coltekin-rama-2016-discriminating, medvedeva-etal-2017-sparse}, and recurrent networks~\cite{guggilla-2016-discrimination, ccoltekin2018tubingen}.

In the specific case of Portuguese, older studies have relied on N-gram-based techniques to achieve results above $90\%$ accuracy on silver-labeled benchmarks~\cite{da2006identification, zampieri2012automatic, goutte2014nrc, malmasi-dras-2015-language, castro2016discriminating}. However, it has been noted that evaluating on silver-labeled corpora is reliability~\cite{zampieri2014varclass}, and  preliminary results obtained on the gold-labeled DSL-TL corpus~\cite{zampieri2023language} revealed more modest performance, with $F_1$ scores below $70\%$. Additionally, contrary to observations in silver-labeled evaluations~\cite{medvedeva-etal-2017-sparse}, the current state-of-the-art result for Portuguese LVI on the DSL-TL benchmark ($0.79~F_1$ score) comes from fine-tuning a collection of BERT-based models~\cite{vaidya2023two}.

\subsection{Cross-Domain Capabilities}
\label{subsec:Delexicalization}

\citet{lui2011cross} revealed that N-grams based techniques had limited cross-domain capabilities for the language identification task. Despite the good results of N-gram-based models when the train and test domain overlap (above 85\% accuracy), the results also show that the effectiveness decreased as much as 40\% when both sets do not match. In order to address this phenomenon, the authors have devised a feature selection mechanism that later opened the door to the development of the first cross-domain language identification tool, the \texttt{langid.py}~\cite{lui2012langid}.

In the context of French LVI,~\citet{diwersy2014weakly} used unsupervised learning to demonstrate that, despite the good results reported by N-grams based-methods (above 95\% accuracy), the feature learned by these models reveal no interest from a linguistic point of view. Instead, classifiers relied on named entities, polarity and thematics embedded in the training corpus to support its inference process (Ex: If ``Cameroun'' was mentioned in the document, the model assigned a French-Cameroonian label to it). 

In light of these facts, the mass adoption of these architectures in the context of LVI, creates urgency for finding solutions to surpass this limitation. In this study, we extend the knowledge about the cross-domain capabilities of N-gram-based models, while presenting the first results for transformer architectures.

%% file: src/Dataset.tex
\section{PtBrVarId Dataset}
\label{sec:create_corpus}

In this section we introduce the PtBrVarId, the first silver-labeled multi-domain Portuguese LVI corpus. This resource resulted from the compilation of open-license corpora from $11$  European (EP) and Brazilian (BP) sources over six domains, namely: \textbf{Journalistic}, \textbf{Legal}, \textbf{Politics}, \textbf{Web}, \textbf{Social Media} and \textbf{Literature}. The following sections describe how the dataset was created.

\input{tables/data}

\subsection{Corpora Compiled}
\label{subsec:compiling}

Training machine learning and deep learning models requires a robust and well-labeled training corpus. However, manually labeling such a corpus is often laborious, time-consuming and expensive. To address this challenge in our research, we opted for a silver labeling approach. 

In the context of the VID task, silver labeling involves identifying texts where the variety can be inferred with a reasonable degree of confidence based on the  documents metadata. In the following paragraphs we describe the data sources used in each textual domain along with the heuristics that supported the silver-labelling step. It is important to note that we were careful to only use sources that were permissive for academic research.

\begin{description}

    \item[Journalistic] As a source of news corpus we use two resources available at Linguateca~\cite{santos2014corpora}, namely: CETEMPublico~\cite{rocha2000cetempublico} and CETEMFolha. The CETEMPublico corpus contains news articles from the Portuguese newspaper ``Público'' while the CETEMFolha contains news from the Brazilian newspaper ``Folha de São Paulo''. The geographic location of the newspaper is used to label the Portuguese variety.
    
    \item[Literature] The literature domain relies on three data sources that index classics of Portuguese literature: the Gutenberg project\footnote{\url{https://www.gutenberg.org/browse/languages/pt\#a4827}}; the LT-Corpus~\cite{genereux2012large}; and the Brazilian literature corpus\footnote{\url{https://www.kaggle.com/datasets/rtatman/brazilian-portuguese-literature-corpus}}. The author's nationality was used  to label the documents as European or BP.

    \item[Legal] The Brazilian split from the legal corpora was compiled from RulingBR~\cite{feijo2018rulingbr} which  contains decisions from the Brazilian supreme court (``Supremo Tribunal Federal'') between 2011 to 2018. The European split was built from the DGSI website\footnote{\url{https://www.dgsi.pt}} which provides access to a set of databases of precedents and to the bibliographic reference libraries of the Portuguese Ministry of Justice.

    \item[Politics] For the politics domain we used the manual transcriptions of political speeches in both the European Parliament~\cite{koehn2005europarl} and the Brazilian Senate~\cite{DVN/M9UU09_2020}. The document's origin was used to infer the label for the Portuguese variety.   

    \item[Web] For the web domain, corpora were extracted from OSCAR~\cite{OrtizSuarezSagotRomary2019}. To define the labels, we began by identifying domains ending in \texttt{.pt} or \texttt{.br}. From this list, we manually curated a set of the $50$ most frequent domains ending in \texttt{.pt} and 50 domains ending in \texttt{.br}. The documents from OSCAR associated with these curated domains were then used in our corpus.
    
    \item[Social Media] The social media corpora derives from three data sources. For BP we used the Hate-BR~\cite{vargas-etal-2022-hatebr} dataset, which was manually annotated for train hate speech classifiers, and a compilation of fake news spread in Brazilian WhatsApp groups~\cite{cunha2021fakewhatsapp}. Regarding EP, the tweets collected by~\citet{ramalho2021highlevel} were filtered based on the tweets' metadata location. Tweets whose location is not part of Wikipedia's list of Portuguese cities\footnote{\url{https://en.wikipedia.org/wiki/List_of_towns_in_Portugal}}, were discarded.
    
\end{description}

Despite the dataset proposed being silver-labeled, some of their components are extracted from high-quality manually annotated corpora that offer sufficient guarantees of belonging to a single language variety. For example, the Europarl corpus~\cite{koehn2005europarl}, is composed of manual transcriptions in EP of political speeches made in European Parliament, therefore it is very unlikely to find any marks of BP in such corpus.

\subsection{Data Cleaning}

To reduce noise in the corpus, we implemented a dedicated data cleaning pipeline. The process starts with basic operations to remove null, empty, and duplicate entries. We then employ the \texttt{clean-text} tool\footnote{\url{https://github.com/jfilter/clean-text}} to correct Unicode errors and standardize the text to ASCII format. For the Web domain, an additional step is taken using the \texttt{jusText} Python package\footnote{\url{https://github.com/miso-belica/jusText}} to filter out irrelevant sentences and remove boilerplate HTML code. Finally, outliers within each domain are identified and removed based on the interquartile range (IQR) of token counts, calculated using the \texttt{nltk} word tokenizer for Portuguese\footnote{\url{https://www.nltk.org/api/nltk.tokenize.word_tokenize.html}}. Texts falling below the first quartile minus 1.5 times the IQR, or above the third quartile plus 1.5 times the IQR, are discarded. This approach effectively eliminates documents that are either too short or too long for their respective domains.

Table~\ref{tab:data_statistics} presents the statistics for the corpus obtained after applying the filtering pipeline. The final corpus comprises 7,304,438 documents, predominantly from the EP segments of the Journalistic, Legal, and Social Media domains. Regarding the number of tokens, we observe that, with the exception of the Journalistic domain, the distribution between documents labeled as EP and BP within each domain is similar.

A comparison across the domains reveals that the Web domain contains the highest average number of tokens per document, whereas the Social Media domain has the lowest, averaging around 18 tokens per document. This disparity is significant for the development of variety identification models, as distinguishing between language varieties in shorter texts is more challenging due to the limited linguistic cues available. Therefore, the Social Media domain is expected to pose more difficulties than the Web domain, where longer texts provide more opportunities to identify distinguishing features of EP and BP.

It is also important to note that the dataset is highly unbalanced across all domains except the Web domain. This imbalance should be carefully considered when training models using this dataset to ensure robust and unbiased effectiveness.

\subsection{Quality Assurance}

To ensure the quality of the silver-labeling process, we asked three linguists to manually annotate 300 documents, focusing on two key aspects:

\begin{description}

\item \textbf{Variety} The linguists were asked to determine the variety of the text. They had three options: EP, BP, or ``Undetermined'' for cases where no variety-specific linguistic features were available.

\item \textbf{Domain} The linguists were also tasked with identifying the domain to which each sentence belonged. They could choose from the six domains used in this research, or select ``Undetermined'' if the domain could not be clearly identified.

\end{description}

For the sampling process, we randomly selected 50 documents from each domain in our corpus, with an equal split of 25 documents silver-labeled as EP and 25 as BP. 

Table~\ref{tab:iaa} presents the agreement between the three annotators using three metrics:

\begin{description}

\item \textbf{Fleiss's Kappa}~\cite{fleiss1971measuring}: Measures the agreement between annotators beyond chance, with values ranging from 0 (no agreement) to 1 (perfect agreement).

\item \textbf{Majority Rate}: Indicates the percentage of texts where two out of three annotators agree on an annotation.

\item \textbf{Accuracy}: Assesses how often the majority vote between annotators matches the automatic annotation. It is important to remark that the cases where the labeled agreed by the annotators is ``Undetermined'' we count both silver-labels (EP and BP) as correct since the text is in fact valid in both varieties.

\end{description}

\input{tables/iaa}

The results obtained show that the agreement is higher for the textual domain aspect than for the language variety. However, the variety aspect still achieves a Fleiss' Kappa of 57\%, which, for three annotators with three labels, can be considered moderate agreement. Upon closer inspection of the results, we found that the Fleiss' Kappa is lower in the Literature, Social Media, and Legal domains (see Table~\ref{tab:annotations_detailed}). For the Social Media domain, we found the disagreement to be mainly driven by the short length of the texts, with ``Undetermined'' representing 42\% of the labels the annotators agreed on. The same was found for the legal domain, which has the second lowest average tokens per document, where the ``Undetermined'' represents 34\% of the labels the annotators agreed on. In the Literature domain, the disagreement is mainly attributed to the corpus consisting of contemporary books, which often blend linguistic features from both European and BP, making it difficult to assign a definitive variety label.

In Table~\ref{tab:annotations_detailed} we detail the annotation agreement metrics per domain for the manually label subset of the PtBrVId corpus. The table shows statistics for the Fleiss' Kappa with all the labels and the Fleiss' Kappa when the entries for which one of the annotators marked the entry as ``Undetermined''. To complete the table we also show the percentage of entries for which at least one annotator labeled as ``Undetermined''.  

\input{tables/iaa_domain}

Nevertheless, a majority consensus among the annotators is almost always achievable (over 90\% of the times) in both aspects. Furthermore, this majority is strongly aligned with the automatic annotations, with agreement between the annotators and the silver labels exceeding 75\%.

In addition to releasing the full annotations provided by each annotator, the documents for which a majority vote could be determined are included in the \textit{test} partition of our dataset. For documents labeled as ``Undetermined'' by the annotators, the original silver label was used as the final label. The complete dataset is publicly accessible on HuggingFace\footnote{\url{https://huggingface.co/datasets/liaad/PtBrVId}}. This dataset offers an opportunity for an in-depth study of the cross-domain capabilities of various LVI techniques, with a particular focus on the application of pretrained transformers, which is the main focus of this paper.

%% file: tables/data.tex
\begin{table*}
\centering
\begin{tblr}{
  colspec = {X[4,l]X[2,c]X[2,si={table-format=2},c]X[2,si={table-format=4},c]X[2,si={table-format=3.2},c]X[2,si={table-format=3.2},c]X[4,c]X[4,c]},
  width = \textwidth,
  column{4,7,8} = {r},
  row{1,2} = {c},
  cell{1}{2} = {r=2}{},
  cell{1}{3} = {c=5}{},
  cell{3}{1} = {r=2}{},
  cell{5}{1} = {r=2}{},
  cell{7}{1} = {r=2}{},
  cell{9}{1} = {r=2}{},
  cell{11}{1} = {r=2}{},
  cell{13}{1} = {r=2}{},
  hline{1,3,15} = {-}{0.08em},
  hline{5,7,9,11,13} = {-}{},
}
             & \textbf{Label} & {{{\textbf{Tokens}}}}       &           &           &            &                & {{{\textbf{Docs}}}} \\
             &                & {{{\textbf{\textbf{Min}}}}} & {{{\textbf{Max}}}} & {{{\textbf{Avg}}}} & {{{\textbf{Std}}}}  & {{{\textbf{Count}}}}    & {{{\textbf{Count}}}}         \\
Journalistic & \includegraphics[width=0.5cm]{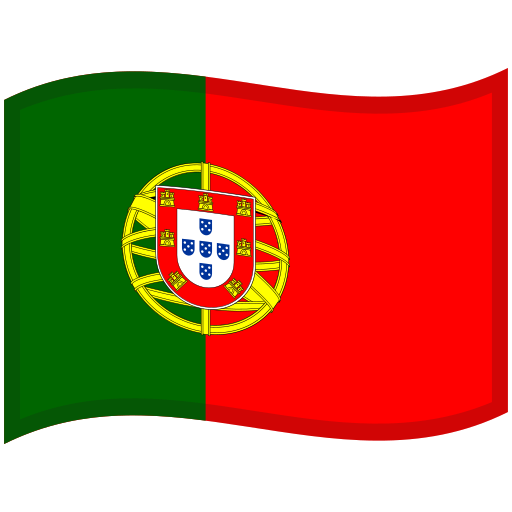}        & 16                          & 475       & 131.29    &  61.45     & 189,506,320      & 1,443,422             \\
             & \includegraphics[width=0.5cm]{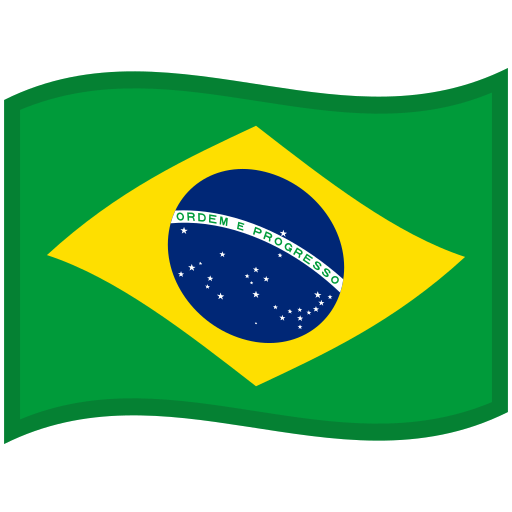}         & 18                          & 560       & 81.09     &  39.11     & 27,077,538       & 333,903              \\
Literature   & \includegraphics[width=0.5cm]{assets/portugal.png}           & 16                          & 186       & 77.20     &  37.39     & 1,859,660        & 24,090               \\
             & \includegraphics[width=0.5cm]{assets/brazil.png}           & 17                          & 185       & 72.55     &  36.19     & 3,805,896        & 52,458               \\
Legal        & \includegraphics[width=0.5cm]{assets/portugal.png}           & 16                          & 139       & 51.63     &  24.43     & 152,717,737      & 2,957,980             \\
             & \includegraphics[width=0.5cm]{assets/brazil.png}           & 20                          & 124       & 47.53     &  22.11     & 221,167         & 4,653                \\
Politics     & \includegraphics[width=0.5cm]{assets/portugal.png}           & 20                          & 798       & 258.32    &  173.39    & 7,203,739        & 27,887               \\
             & \includegraphics[width=0.5cm]{assets/brazil.png}           & 21                          & 796       & 276.97    &  177.60    & 1,012,586        & 3,656                \\
Web          & \includegraphics[width=0.5cm]{assets/portugal.png}           & 22                          & 2042      & 517.96    &  414.72    & 22,598,587       & 43,630               \\
             & \includegraphics[width=0.5cm]{assets/brazil.png}           & 15                          & 2075      & 539.66    &  463.16    & 23,913,771       & 44,313               \\
Social Media & \includegraphics[width=0.5cm]{assets/portugal.png}           & 3                           & 646       & 18.94     &  9.85      & 44,758,304       & 2,363,261             \\
             & \includegraphics[width=0.5cm]{assets/brazil.png}           & 6                           & 51        & 17.11     &  10.17     & 94,177          & 5,504                
\end{tblr}
\caption{Summary statistics of the PtBrVarId corpus, including the minimum, maximum, average, standard deviation, and count of tokens, as well as the number of documents for each domain and label.}
\label{tab:data_statistics}
\end{table*}

%% file: tables/iaa.tex
\begin{table}
\centering
\begin{tblr}{
  colspec = {X[2,l]X[3,l]X[3,c]},
  width = \columnwidth,
  row{1} = {c},
  cell{2}{1} = {r=3}{},
  cell{5}{1} = {r=3}{},
  hline{1,2,8} = {-}{0.08em},
  hline{5} = {-}{0.05em},
}
        & \textbf{Metric} & \textbf{Result} \\
\textbf{Variety} & Fleiss' Kappa   & 0.57                 \\
        & Majority Rate   & 0.95                 \\
        & Accuracy        & 0.86                 \\
\textbf{Domain}  & Fleiss' Kappa   & 0.69                 \\
        & Majority Rate   & 0.94                 \\
        & Accuracy        & 0.76                 
\end{tblr}
\caption{Agreement among the three annotators regarding  language variety and textual domain.}
\label{tab:iaa}
\end{table}

%% file: tables/iaa_domain.tex
\begin{table}[ht]
\centering
\begin{tblr}{
  colspec = {X[3,l]X[4,l]X[3,c]},
  width = \columnwidth,
  row{1} = {c},
  cell{2}{1} = {r=3}{},
  cell{5}{1} = {r=3}{},
  cell{8}{1} = {r=3}{},
  cell{11}{1} = {r=3}{},
  cell{14}{1} = {r=3}{},
  cell{17}{1} = {r=3}{},
  hline{1,20} = {-}{0.08em},
  hline{2,5,8,11,14,17} = {-}{0.05em},
}
\textbf{Domain} & \textbf{Metric}       & \textbf{Result} \\
Literature      & Fleiss' Kappa          & 0.23            \\
                & Fleiss' Kappa$_{wo/u}$ & 0.51            \\
                & Undetermined Rate             & 0.36            \\
Legal           & Fleiss' Kappa          & 0.46            \\
                & Fleiss' Kappa$_{wo/u}$ & 0.73            \\
                & Undetermined Rate             & 0.34            \\
Politics        & Fleiss' Kappa          & 0.78            \\
                & Fleiss' Kappa$_{wo/u}$ & 0.87            \\
                & Undetermined Rate             & 0.10            \\
Web             & Fleiss' Kappa          & 0.67            \\
                & Fleiss' Kappa$_{wo/u}$ & 0.84            \\
                & Undetermined Rate             & 0.20            \\
Social Media    & Fleiss' Kappa          & 0.53            \\
                & Fleiss' Kappa$_{wo/u}$ & 0.94            \\
                & Undetermined Rate             & 0.42            \\
Journalistic    & Fleiss' Kappa          & 0.72            \\
                & Fleiss' Kappa$_{wo/u}$ & 0.90            \\
                & Undetermined Rate             & 0.04             
\end{tblr}
\caption{Extended per-domain analysis of annotator agreement. We present Fleiss' Kappa  for all three labels, as well as Fleiss' Kappa excluding the ``Undetermined'' documents (Fleiss' Kappa$_{wo/u}$). The ``Undetermined Rate'' rows shows the percentage of documents for which at least one annotator labeled as ``Undetermined''.  }
\label{tab:annotations_detailed}
\end{table}

%% file: src/Experimental_Setup.tex
\section{Experimental Setup}
\label{sec:setup}

In this study, we investigate the effectiveness of fine-tuning a transformer-based model for the Portuguese LVI task. We employ an iterative methodology to identify the optimal strategy for combining training corpora from various domains into a unified training process. Our primary objective is to evaluate cross-domain effectiveness and the generalization capabilities of our models.

\subsection{Models \& Baselines}

For the transformer-based model, we use BERTimbau with 334 million parameters~\cite{souza2020bertimbau}. BERTimbau is the result from fine-tuning the original BERT model~\cite{devlin2019bertpretrainingdeepbidirectional} on a Portuguese corpus.

To establish a baseline for comparison with the BERT model, we employ N-grams combined with Naive Bayes classifiers. This choice is motivated by the proven effectiveness of such models in previous LVI studies across various Indo-European languages, including Portuguese~\cite{zampieri2012automatic}.

\subsection{Cross-Domain Training Protocol}
To ensure that our model generalizes effectively across different domains, we define a two-step training protocol. Step one is used to find the best hyperparameters to train the model so ensure the generalization capability of the model. In this step,  the model is trained on a single domain from the PtBrVid corpus and validated on the remaining domains (excluding the one used for training). The hyperparameters yielding the best performance in this cross-domain validation are then used in step two to train the model across all domains combined.

Delexicalization of the corpus is treated as a hyperparameter in our approach. We adjust the probabilities of replacing tokens found by Named Entity Recognition (NER) and Part-of-Speech (POS) tagging with the generic label (such as \texttt{LOCATION} or \texttt{NOUN}), varying these probabilities incrementally from 0\% to 100\% in 20\% steps. It is important to note that delexicalization is applied exclusively to the training set. The validation set remains unaltered, simulating a real-world scenario where the input text is not modified. We leave the study of the impact of delexicalizing the validation set on the effectiveness of the model for future research.

\subsection{Train \& Validation Data}
As referred above the PtBrVId dataset is used to train the models. However, before using for the training, we leave 1,000 documents of each domain for the validation of the model, 500 of each label. 

In the step one of our training protocol, we use 8,000 documents from each domain (4,000 from each label) to train the models. We found this sample size to be enough for the models to converge and ensure fast iteration in the training process.

For step two of our training protocol, we compile all the documents from the PtBrVid corpus including the ones used for validation in step one. To avoid the training being dominated by the more represented domains, we undersample the dataset so that all labels from all domains are equally represented. At this step, the manually annotated set from PtBrVId set is used to keep track of the generalization loss.

\subsection{Benchmarks}

In our evaluation, we use two benchmarks: the DSL-TL and FRMT datasets. As mentioned above, the DSL-TL dataset is the standard benchmark for distinguishing between EP and BP, annotated with three labels: ``EP'', ``BP'', and ``Both''. For our purposes, we exclude documents labeled ``Both'' since our training corpus does not contain that label. This results in a test set comprising 588 documents for BP and 269 for EP. The FRMT dataset~\cite{riley2022frmt} has been manually annotated to evaluate variety-specific translation systems and includes translations in both EP and BP. We adapt this corpus for the VID task, resulting in a dataset containing 5,226 documents, with 2,614 labeled as EP and 2,612 as BP.

\section{Implementation Details}
\label{app:hyper}

NER and POS tags were identified using spaCy\footnote{\url{https://spacy.io/models/pt}}. The BERT model was trained with the \texttt{transformers}\footnote{\url{https://huggingface.co/docs/transformers/}} and \texttt{pytorch}\footnote{\url{https://pytorch.org}} libraries, for a maximum of 30 epochs, using early stopping with a patience of three epochs, binary cross-entropy loss, and the AdamW optimizer. The learning rate was set to $2 \times 10^{-5}$. In addition, a learning rate scheduler was used to reduce the learning rate by a factor of 0.1 if the training loss did not improve for two consecutive epochs. N-gram models were trained using the \texttt{scikit-learn}\footnote{\url{https://scikit-learn.org/}} library. The following hyperparameters were taken into account in the grid search we performed"

\begin{itemize}
    \item \textbf{TF-IDF Max Features:} The number of maximum features extracted using TF-IDF was tested with the following values: 100, 500, 1,000, 5,000, 10,000, 50,000, and 100,000.
    \item \textbf{TF-IDF N-Grams Range:} The range of n-grams used in the TF-IDF was explored with the following configurations: (1,1), (1,2), (1,3), (1,4), (1,5), and (1,10).
    \item \textbf{TF-IDF Lower Case:} The effect of case sensitivity was tested, with the lowercasing of text being either \texttt{True} or \texttt{False}.
    \item \textbf{TF-IDF Analyzer:} The type of analyzer applied in the TF-IDF process was either \texttt{Word} or \texttt{Char}.
\end{itemize}

Regarding computational resources, this study relied on Google Cloud N1 Compute Engines to perform the tuning and training of both the baseline and the BERT architecture. For the baseline, an N1 instance with 192 CPU cores and 1024 GB of RAM was used. For BERT, we used an instance with 16 CPU cores, 30 GB of RAM, and 4x Tesla T4 GPUs. The grid search on N-grams takes approximately three hours under these conditions, while for BERT, it takes approximately 52 hours to complete. The final training took three hours for N-grams and approximately ten hours for BERT.

We have made our codebase open-source\footnote{\url{https://github.com/LIAAD/portuguese_vid}} to promote reproducibility of our results and to encourage further research in this area. 

%% file: src/Results.tex
\section{Results}
\label{sec:results}

\subsection{Impact of Delexicalization}
\label{subsec:hyper}

\begin{figure}[ht]
    \centering
    \includegraphics[width=0.8\columnwidth]{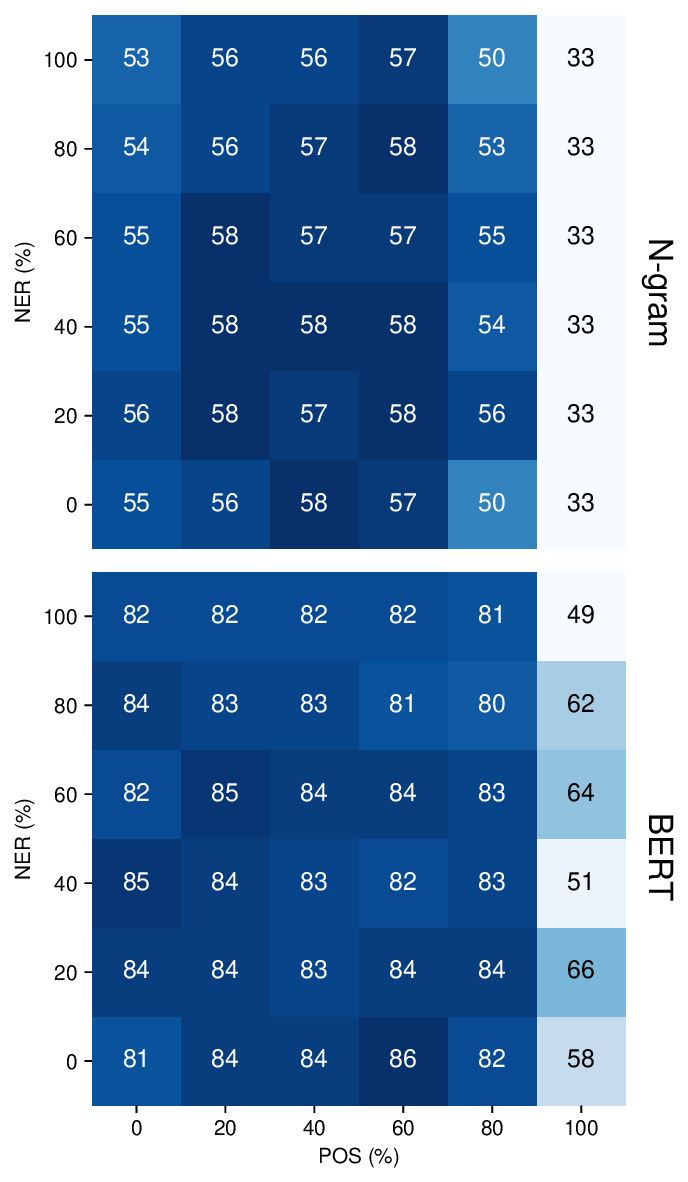}
    \caption{Average $F_1$ score for each ($P_\text{POS}$, $P_\text{NER}$).}
    \label{fig:hyperparameters_ngrams}
\end{figure}

Figure~\ref{fig:hyperparameters_ngrams} depicts the average $F_1$ scores obtained in the PtBrVid validation set by the N-grams and BERT models, for each ($P_\text{POS}$, $P_\text{NER}$) percentage pair. The averages are computed across models trained in different domains.

The results suggest that intermediate levels of delexicalization can yield marginal improvements in model effectiveness. However, high levels of $P_\text{POS}$ adversely affect model performance. This finding is particularly interesting because previous studies have reported significant reductions in effectiveness due to delexicalization~\cite{sharoff2010web,lui2014exploring}. Notably, these earlier studies focused solely on full delexicalization and did not evaluate performance on out-of-domain corpora.

Based on these insights, we proceeded to the second step of our training protocol using a delexicalized version of the training set, with ($P_\text{POS}=0.2$, $P_\text{NER}=0.6$) for the N-gram model and ($P_\text{POS}=0.6$, $P_\text{NER}=0.0$) for BERT models.

\subsection{Overall Results}

This section presents the $F_1$ scores for the N-gram baseline and BERT fine-tuning models, comparing their performance with and without delexicalization to highlight their impact on the overall effectiveness of the model.

\begin{figure}
    \centering
    \includegraphics[width=\columnwidth]{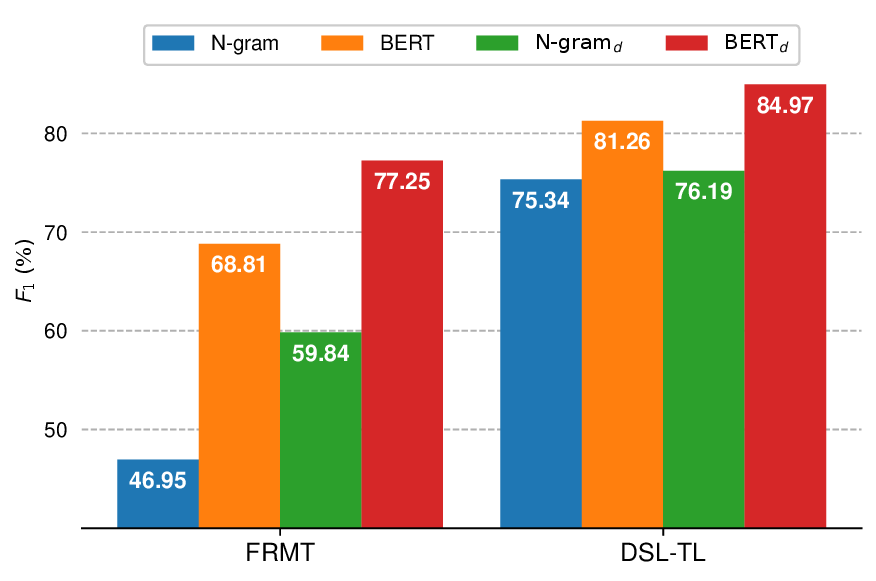}
    \caption{$F_1$ in FRMT and DSL-TL benchmarks. Models with the subscript $d$ were trained on a delexicalized corpus.}
    \label{fig:results}
\end{figure}

The results in Figure~\ref{fig:results} underline the benefits of delexicalization on system effectiveness across both benchmarks and models. Specifically, in FRMT, training in the delexicalized corpus improved the $F_1$ score by approximately 13 and 10 percentage points for the N-gram and BERT models, respectively. 

Upon examining the less pronounced discrepancy in the DSL-TL benchmark, we found it to be largely attributed to the FRMT dataset's entity-specific partition, known as the entity bucket. In this bucket, models trained without delexicalization struggle, as they rely on entities to determine language variety. Given that the FRMT dataset contains the same text in both BP and EP, these models often misclassify pairs of sentences by assigning the same label to both, leading to frequent errors. In the extreme case, they end up getting around half of the labels wrong, which is what happened to the N-gram model, only achieving an $F_1$ score of 46.95\% in this benchmark. This highlights the importance of using delexicalization in the training process. To the best of our knowledge, we are the first to report positive results from the use of delexicalization, which was enabled by the proposed cross-domain training protocol. 

When comparing the BERT model with the N-gram models, one can observe that the BERT model outperforms the N-gram model across all scenarios, achieving an $F_1$ score of  84.97\% in DSL-TL and 77.25\% in FRMT. To support further research and exploration, we have made the BERT$_d$ model available on HuggingFace, inviting the research community to use and build on this work\footnote{\url{https://huggingface.co/liaad/PtVId}}.

%% file: src/Conclusion.tex
\section{Conclusion \& Future Work}
\label{sec:conclusions}

In this study, we introduced the first multi-domain Portuguese LVI corpus, which includes more than 7 million documents. Leveraging this corpus, we fine-tuned a BERT-based model to create a robust tool for discriminating between European and Brazilian Portuguese. The training strategy leverages delexicalization to mask entities and thematic content in the training set, thereby enhancing the model's ability to generalize. This approach has potential for adaptation to other language variants and languages.

We have identified two key avenues for future work to further enhance the quality and scope of Portuguese LVI. First, the corpus should be expanded to include other less-resourced Portuguese varieties, particularly African Portuguese. Second, it is crucial to explore the impact of the pre-trained model selection, as the language variety on which the model was originally trained may introduce bias into the LVI classifier.

%% file: src/aknowledments.tex
\section*{Acknowledgments}

This research is supported by national funding from the Portuguese Foundation for Science and Technology (FCT) under the project with DOI \texttt{\small 10.54499/LA/P/0063/2020}. The authors also acknowledge the support of the StorySense project (DOI \texttt{\small 10.54499/2022.09312.PTDC}) and the advanced computing project PTicola (ID \texttt{\small CPCA-IAC/AV/594794/2023}). Hugo Sousa further acknowledges FCT for funding his PhD grant (ID \texttt{\small 2022.14691.BD}).